\documentclass[journal]{IEEEtran}

\usepackage{graphicx}
\usepackage{amsmath}
\usepackage{amssymb}
\usepackage{xcolor}
\usepackage{verbatim}
\usepackage{hyperref}
\usepackage{algorithm}
\usepackage{algpseudocode}
\usepackage{listings}
\usepackage{caption}
\usepackage{subcaption}
\usepackage{authblk}
\usepackage{gensymb}

\usepackage[noadjust]{cite}
\usepackage[labelformat=simple]{subcaption}

\begin{document}
\title{SQ-Swin: a Pretrained Siamese Quadratic Swin Transformer for Lettuce Browning Prediction}

\author{Dayang Wang$^1$, 
        Boce Zhang$^2$,
        Yongshun Xu$^1$,\\
        Yaguang Luo$^{3*}$, 
        Hengyong Yu$^{1^*}$, \textit{IEEE Senior Member}
\thanks{* Drs. Yaguang Luo and Hengyong Yu serve as co-corresponding authors. }
\thanks{
$^1$Dayang Wang, Yongshun Xu, and Hengyong Yu are with Department of Electrical and Computer Engineering, University of Massachusetts, Lowell, MA, USA}
\thanks{$^2$Boce Zhang is with Department of Food Science and Human Nutrition, University of Florida, Gainesville, FL, USA}
\thanks{$^3$Yaguang Luo is with Food Quality Lab / Environmental Microbial and Food Safety Lab, Agricultural Research Service, United States Department of Agriculture, Beltsville, MD, 20705, US}
}

\markboth{IEEE }%
{Shell \MakeLowercase{\textit{et al.}}: Bare Demo of IEEEtran.cls for IEEE Journals}

\maketitle

\begin{abstract}
Enzymatic browning is a major quality defect of packaged “ready-to-eat” fresh-cut lettuce salads. While there have been many research and breeding efforts to counter this problem, progress is hindered by the lack of a technology to identify and quantify browning rapidly, objectively, and reliably. Here, we report a deep learning model for lettuce browning prediction. To the best of our knowledge, it is the first-of-its-kind on deep learning for lettuce browning prediction using a pretrained Siamese Quadratic Swin (SQ-Swin) transformer with several highlights. First, our model includes quadratic features in the transformer model which is more powerful to incorporate real world representations than the linear transformer. Second, a multi-scale training strategy is proposed to augment the data and explore more of the inherent self-similarity of the lettuce images. Third, the proposed model uses a siamese architecture which learns the inter-relations among the limited training samples. Fourth, the model is pretrained on the ImageNet and then trained with the reptile meta-learning algorithm to learn higher-order gradients than a regular one. Experiment results on the fresh-cut lettuce datasets show that the proposed SQ-Swin outperforms the traditional methods and other deep learning-based backbones.
\end{abstract}

\begin{IEEEkeywords}
lettuce, enzymatic browning, transformer, quadratic, siamese model, reptile. 
\end{IEEEkeywords}

\IEEEpeerreviewmaketitle

\section{Introduction}
Lettuce (Lactuca sativa L.) is an important vegetable crop worldwide and a common ingredient in packaged 'ready-to-eat' salads. However, browning discoloration on the leaf ribs' cut edges significantly reduces the vegetable quality and shortens the shelf-life  \cite{he2007enzymatic,zhou2004determination}. Many studies have shown that the browning of fresh-cut lettuce is caused by the enzymatic oxidation of phenolic metabolites \cite{taranto2017polyphenol}. The tissue injuries sustained during cutting disrupt the natural compartmentation of polyphenol oxidase and phenolics, thus accelerating the browning process \cite{cantos2001effect,luna2016modified}. 

Developing technologies to accurately identify, quantify, and predict browning is critical, yet, achieving both reliability and efficiency is challenging. Commonly used colorimeters are useless since the cut edges are much narrower than the available smallest probes. Visual evaluation approaches are not only time consuming but also subjective. Digital imaging following feature extracting with RGB or Lab improved objectivity and speed over the traditional visual method. However, multiple values required to depict the discoloration made it impossible for data comparison among lettuce cultivars. 



Compared to the traditional learning models, deep learning models can be trained end-to-end to learn the high-dimensional features incrementally, thereby eliminating the need for domain expertise and manual feature extraction. Due to the advantages, a plethora of deep learning algorithms have been developed in many science and engineering fields \cite{he2016deep,fan2021sparse,wu2020md,li2020human,wang2022manifoldron,li2020hrnet,jia2021nondestructive}. Very recently, the transformer has been featured as the most popular backbone architecture \cite{RN58,RN59,RN62,RN67,chen2021transmorph,gu2022sthardnet,he2022transformers,shamshad2022transformers,xiong2022swin,wu2021elnet}. Dosovitskiy \textit{et al.} first introduced the transformer to the computer vision (CV) field by mapping a 16×16 image patch into a word sequence \cite{RN56}. Based on the first-of-its-kind vision transformer, Yuan \textit{et al.} introduced a Token-to-token process to further enrich the token selection from the image. Wang \textit{et al.} further extended the Token-to-Token method to low-dose CT denoising and obtained more perceptually pleasing CT images \cite{wang2022ctformer}. Touvron \textit{et al.} introduced the distillation to the transformer by taking the convnet as a teacher and the transformer as a student, thus achieving comparable performance with convnet on ImageNet \cite{deng2009imagenet}. Han \textit{et al.} designed a transformer by considering the attention inside the image patches. Liu \textit{et al.} further employed hierarchical shifted window attention on the vision transformer and achieved the state-of-the-art performances in image classification, object detection, and semantic segmentation \cite{RN74}. 

The transformer models have proved their successes in various computer vision tasks. However, their input tokens fed into the inherent self-attention module are mapped from the image features using the simple linear neuron. Nevertheless, the bio-neuron is far more complex and diverse in the real world. Quadratic neuron is one of the critical neurons that features quadratic mapping across the neural layers. Research has shown that quadratic neurons encode more rich features than linear features  \cite{fan2019quadratic,sarao2020optimization,fan2020universal,xu2021robust}. Therefore, we are motivated for the first time to introduce the quadratic neuron to the transformer and prototype a high-order transformer based on Swin transformer modules. Naturally, we are curious whether the second-order SQ-Swin can deliver competitive performances in the lettuce browning prediction field, hence, transforming the theoretical deep learning model into proper deep learning applications.

To the best of our knowledge, the contributions of this paper are fourfold:
i) The first deep learning model is developed for lettuce browning prediction. ii) A quadratic transformer model is first proposed to empower the token mapping neurons with more powerful representation ability. iii) The siamese architecture is introduced to the quadratic transformer to learn the inter-relations among the limited samples. Then, the reptile meta-learning algorithm is adopted to train the proposed model. iv) With the merits of transfer learning, the SQ-Swin is pretrained on the ImageNet, then trained and evaluated on the fresh cut lettuce data. Comprehensive experiments demonstrate the high efficacy and the efficiency of the SQ-Swin on the lettuce browning prediction problem.

\section{Related works}
\textbf{Quadratic neuron.}
The mainstream deep neural network uses linear neurons for feature inference. However, in the real world, the bio-neuron such as that in the human brain's nerve cell is fundamentally more diverse and complex. Fan \textit{et al.} prototyped a quadratic neural network by replacing the inner product in the traditional artificial neuron with a quadratic operation \cite{fan2019quadratic}. The quadratic neuron is the first high-order neuron. It has been explored in a plethora of researches and shown the advantage over the traditional linear neuron in representation and efficiency \cite{fan2019quadratic,sarao2020optimization,fan2020universal,xu2021robust}. 

Mathematically, given an input $\boldsymbol{x} \in \mathrm{R}^n$, a quadratic neuron $\mathrm{q}$ is characterized as
\begin{equation}
    \mathrm{q}(\boldsymbol{x}) = \sigma( (\boldsymbol{x}^\top \mathbf{w}^r+\mathbf{b}^r)(\boldsymbol{x}^\top \mathbf{w}^g+\mathbf{b}^g) + (\boldsymbol{x} \odot \boldsymbol{x})^\top \mathbf{w}^b+\mathbf{b}^b ), 
\end{equation}
where $\mathbf{w}^r,\mathbf{w}^g,\mathbf{w}^b \in \mathrm{R}^n$ are weights and $\mathbf{b}^r,\mathbf{b}^g,\mathbf{b}^b \in \mathrm{R}^n$ are biases. $\sigma(\cdot)$ is regular nonlinear activation function such as ReLU. $\odot$ denotes element-wise product. We denote $\mathbf{w^r},\mathbf{b^r}$ as linear terms, and $\mathbf{w}^g$, $\mathbf{b}^g$, $\mathbf{w}^b$, and $\mathbf{b}^b$ as quadratic terms. Notably, that the definition of the quadratic neuron only requires $\mathcal{O}(3n)$ parameters, which is much less than the regular quadratic complexity of $\mathcal{O}(\frac{n(n+1)}{2})$. Furthermore, to facilitate the training of the quadratic parameters and guarantee the model convergence, Fan \textit{et al.} introduced a \textit{Relinear} strategy which includes a special initialization on the quadratic terms:  $\mathbf{w}^g=0$, $\mathbf{b}^g=1$, $\mathbf{w}^b=0$, and $\mathbf{b}^b=0$, assisted by a shrunk small learning rate to prevent magnitude explosion \cite{fan2021expressivity}. 

\textbf{Siamese model.}
The siamese model was proposed to address the few-shot learning problem by learning the inter-relations among images \cite{koch2015siamese}, and it has been explored in various applications \cite{bertinetto2016fully,chicco2021siamese,chen2021exploring,he2018twofold}. Bertinetto \textit{et al.} adopted the fully-convolutional siamese design in video object detection and achieved high performances in many benchmarks \cite{bertinetto2016fully}. Guo \textit{et al.} proposed a simple yet effective siamese fully convolutional classification and regression network for visual tracking and achieved leading performance in real-time speed \cite{guo2020siamcar}. Chen \textit{et al.} developed a siamese transformer to extract the non-local features from the multitemporal image pairs and demonstrated the model's potentials in the multitemporal remote sensing interpretation tasks \cite{chen2022dual}.


\section{Methods}
The lettuce browning prediction task is a typical regression problem. Given data $(\boldsymbol{x},\boldsymbol{y})$, a general deep regression model aims to minimize the mean square error (MSE) between the predictions and the labels: 
\begin{equation}
   \underset{\boldsymbol\theta}{ \min} \ \  \mathcal{J}(\boldsymbol{\theta};\boldsymbol{x}) = \frac{1}{N} \left \| f(\boldsymbol{\theta};\boldsymbol{x})-\boldsymbol{y} \right \|^2,
    \label{define}
\end{equation}
where $f(\boldsymbol{\theta};\boldsymbol{x})$ is the model, $\boldsymbol{\theta}$ is a collection of the model parameters, and $N$ is the number of samples.

To further explore the inter-relations among the limited data, a siamese design is introduced to the proposed SQ-Swin transformer. Given data pairs $(\boldsymbol{x_0},\boldsymbol{y_0})$ and $(\boldsymbol{x_1},\boldsymbol{y_1})$, suppose the two base quadratic swin transformers generate features and predictions ($\boldsymbol{f_0}, \boldsymbol{\hat{y}_0}$) and ($\boldsymbol{f_1}, \boldsymbol{\hat{y}_1}$), respectively. Two losses are used in the SQ-Swin model during optimization: prediction loss and siamese loss. The prediction loss generalizes the margin between the predictions and the true labels. The siamese loss is based on the prior that the linear combination of specific features determines the lettuce browning level. Therefore, the difference between the features should indicate the difference between the labels. Specifically, 
The prediction loss is defined as: 
\begin{equation}
    \mathcal{L}_p = \left \| \boldsymbol{y_0} - \boldsymbol{\hat{y}_0}  \right \|^2 
    + \left \| \boldsymbol{y_1} - \boldsymbol{\hat{y}_1}  \right \|^2.
\end{equation}
The siamese loss is defined as:
\begin{equation}
    \mathcal{L}_s = \left \| \left \| \boldsymbol{f_0} - \boldsymbol{f_1}  \right \|^2 - \left \| \boldsymbol{y_0} - \boldsymbol{y_1}  \right \|^2 \right \|_1. 
\end{equation}
The total loss is the combination of the two losses with a balance parameter $\alpha$,
\begin{equation}
    \mathcal{L} = \mathcal{L}_p + \alpha \cdot \mathcal{L}_s.  
\end{equation}

\begin{figure*}
\centering
\includegraphics[width=\textwidth]{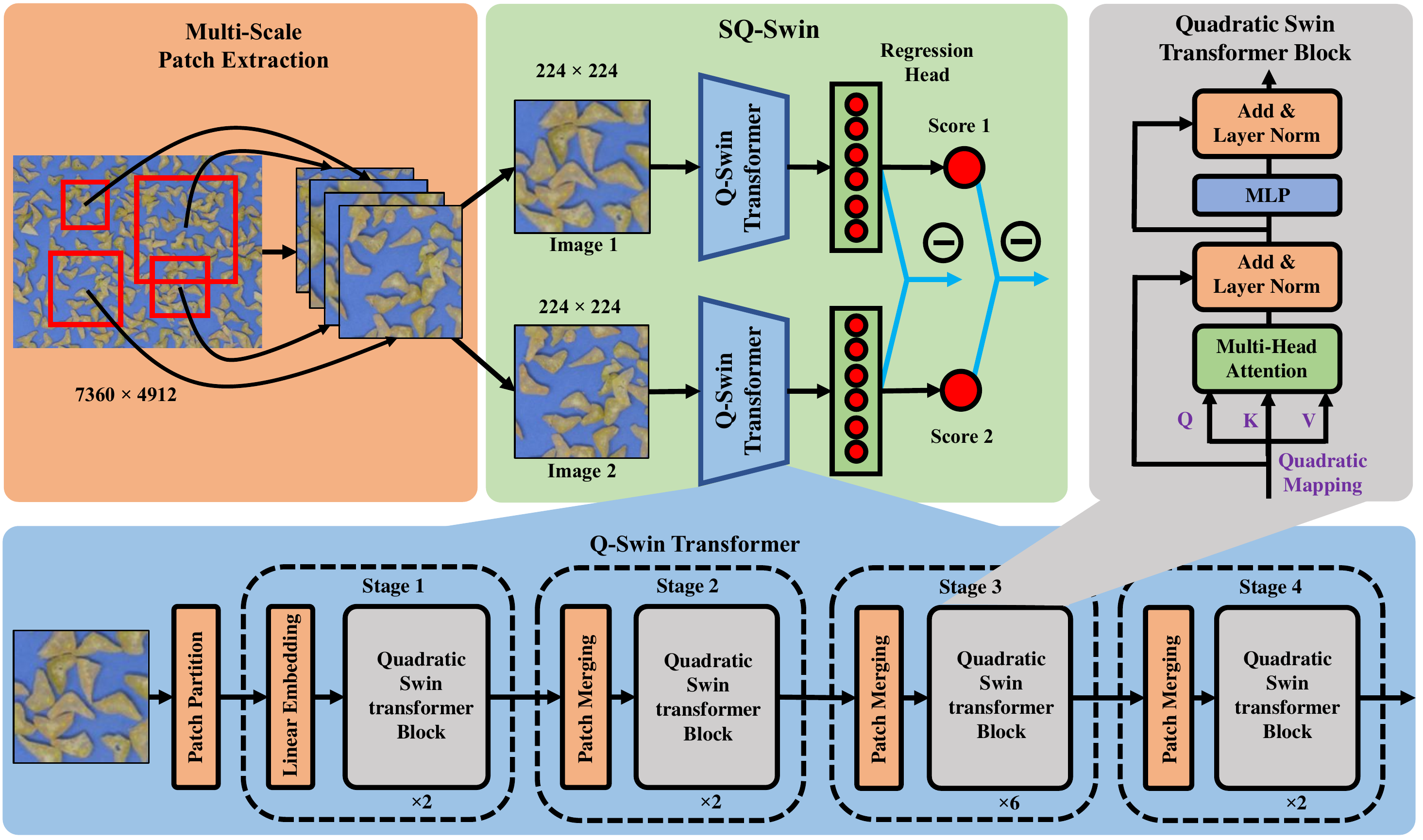}
\caption{The architecture of 
proposed siamese quadratic swin transformer (SQ-Swin).} 
\label{whole}
\end{figure*}

\subsection{Model Architecture}
This paper proposes a siamese quadratic swin transformer for lettuce browning prediction (SQ-Swin). As shown in Fig. \ref{whole}, the multiple-scale patch extraction stage first generates scaled image patches for latter training process. Then, image patches are randomly paired to feed into the SQ-Swin model. Specifically, the SQ-Swin takes the siamese design as the base architecture, which includes two routes of quadratic Swin transformer (Q-Swin) for feature inference. The two Q-Swins share identical designs and parameters. Finally, two multilayer perceptrons (MLPs) are attached on top of the Q-Swin for feature extraction and regression, respectively. 

\begin{figure}
\centering
\includegraphics[width=\linewidth]{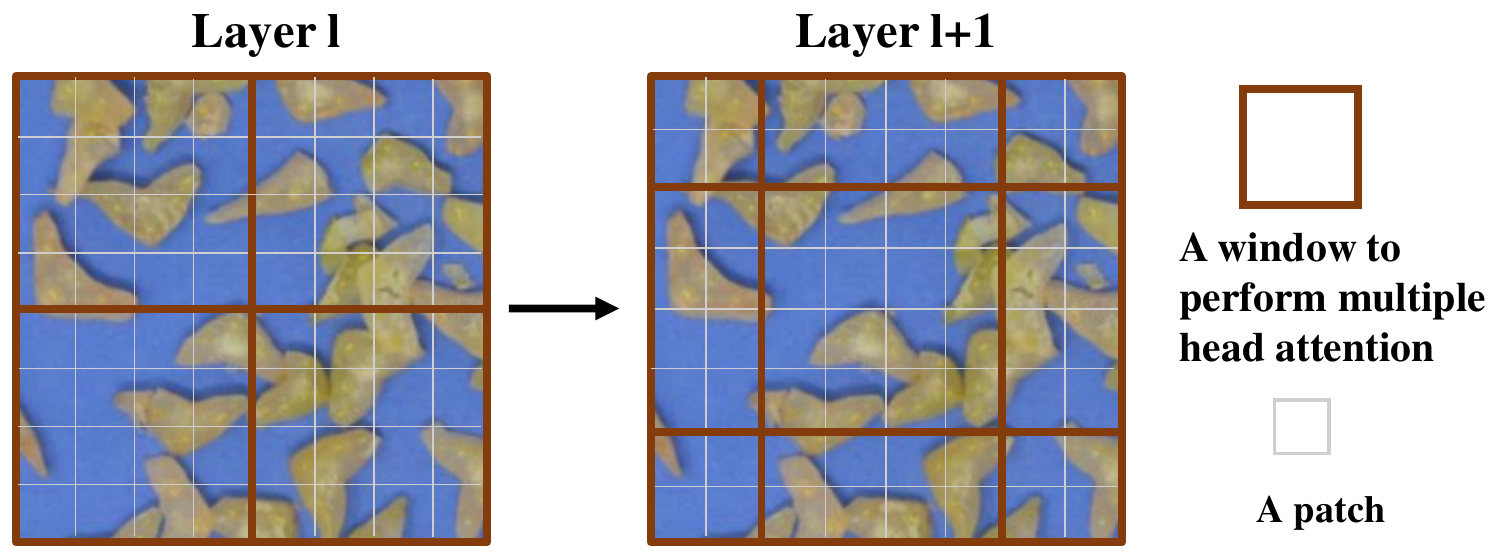}
\caption{Shift window attention.} 
\label{window}
\end{figure}

\textbf{Multi-scale patch extraction.}
Multiple scale visual patterns are essential in semantic segmentation, fine-grained classification, and object detection \cite{lin2017feature,kirillov2019panoptic,gao2019res2net,guo2020augfpn,pauly2003multi}. 
Inspired by these works, we seek to exploit the multi-scale information from the lettuce images. In Contrast to the common methods that incorporate multi-scale features in the network design, we derive the multi-scalability directly from the images by taking advantage of their self-similarities. Specifically, the training images are extracted from multi-scale patches of different sizes from the large raw lettuce image, then resized to a uniformed shape when fed to the models. Thus, the training images are a collections of mixed lettuce images with various feature scales, therefore boosting the generalizability and robustness of the model. 

\textbf{Q-Swin transformer.}
Recently, the Swin transformer has been very popular as a backbone architecture \cite{RN74}. It introduced the hierarchical attention with shifted windows as shown in Fig. \ref{window}. Specifically, given an input image, the patch partition process first splits the image into non-overlapped patches. Then linear embedding module maps these patches into tokens to feed to the transformer blocks. Several blocks with attention are applied for the feature inference. To reduce computational complexity and facilitate hierarchical representation, the patch merging layers between successive transformer blocks are applied to reduce the feature map size by half. Based on the design of the Swin transformer, the proposed quadratic Swin transformer further employs a quadratic transformer block for window attention rather than a regular linear transformer. 

\underline{Quadratic Swin Transformer Block:} Instead of calculating the attention globally where the relations among all the tokens from all the patches are calculated, the swin transformer calculates attention within a certain window as shown in Fig. \ref{window}. In a transformer block, for a given token sequence $\mathbf{T} \in \mathbb{R}^{b \times n \times d}$ within a window, they are first mapped to three matrices of query, key and value $\mathbf{Q},\mathbf{K},\mathbf{V}\in \mathbb{R}^{b \times n \times d}$. Here $b$ is the batch size, $n$ is the number of tokens, and $d$ is the token embedding dimension. The output of the multiple head attention (MHA) is calculated as
\begin{equation}
    \mathrm{MHA}(\mathbf{Q},\mathbf{K},\mathbf{V}) = \mathrm{softmax}(\frac{\mathbf{Q}\mathbf{K^\top}}{\sqrt{d_k}}) \mathbf{V},
\end{equation}
where $\frac{1}{\sqrt{{{d}_{k}}}}$ is the scaling factor determined by the network depth. Instead of using the traditional linear multiple layer perceptron (MLP) to calculate $\mathbf{Q},\mathbf{K}$, and $\mathbf{V}$, the proposed quadratic transformer adopts quadratic multiple layer perceptron (QMLP) to obtain the three matrices, 
\begin{equation}
\begin{cases}
    & \mathbf{Q} = (\mathbf{T} \mathbf{W}_q^r)(\mathbf{T} \mathbf{W}_q^g) + (\mathbf{T} \odot \mathbf{T}) \mathbf{W}_q^b\\
    & \mathbf{K} = (\mathbf{T} \mathbf{W}_k^r)(\mathbf{T} \mathbf{W}_k^g) + (\mathbf{T} \odot \mathbf{T}) \mathbf{W}_k^b\\
    & \mathbf{V} = (\mathbf{T} \mathbf{W}_v^r)(\mathbf{T} \mathbf{W}_v^g) + (\mathbf{T} \odot \mathbf{T}) \mathbf{W}_v^b,\\
\end{cases}
\label{qkv}
\end{equation}

where $\mathbf{W}_q^r$, $\mathbf{W}_q^g$, $\mathbf{W}_q^b$, $\mathbf{W}_k^r$, $\mathbf{W}_k^g$, $\mathbf{W}_k^b$, $\mathbf{W}_v^r$, $\mathbf{W}_v^g$, $\mathbf{W}_v^b \in \mathbb{R}^{d \times d}$ are linear operators. The biases are omited for simplicity. After MHA, layer normalization (LN) and shortcuts are also applied for more feature inference. As shown in Fig. \ref{whole}, the quadratic swin transformer block is characterized as: 
\begin{equation}
\begin{cases}
   & \mathbf{T}^{'}= \mathrm{LN}( \mathrm{MHA}( \mathrm{QMLP} (\mathbf{T}))) + \mathbf{T} \\
   & \mathbf{T}_{\mathrm{q}}=\mathrm{LN}( \mathrm{MLP}(\mathbf{T}')) + \mathbf{T}^{'},
\end{cases}
\end{equation}
where $\mathbf{T'}, \mathbf{T_q}\in \mathbb{R}^{b \times n \times d}$.

\subsection{Model Training}
\begin{figure}
\centering
\includegraphics[width=0.85\linewidth]{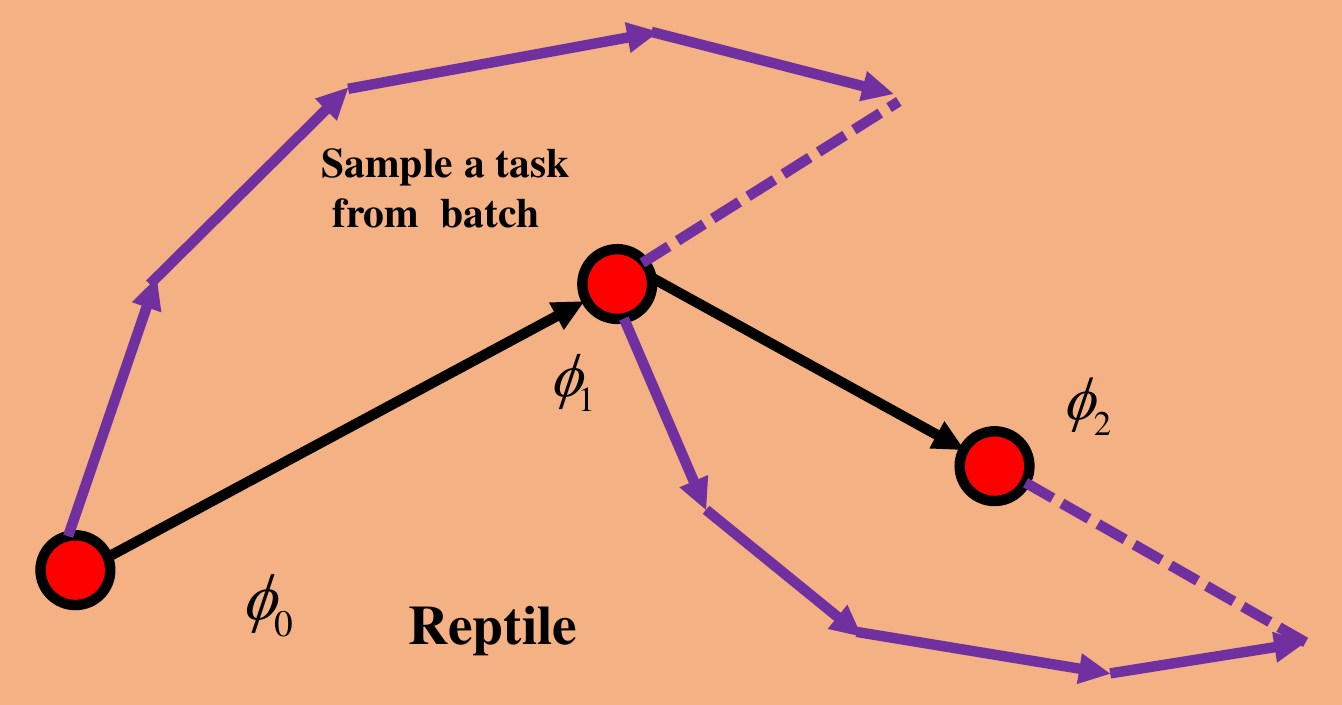}
\caption{The update of the model parameters $\phi$ using reptile.} 
\label{reptile}
\end{figure}

The SQ-Swin can be trained regularly like common deep learning models using optimization algorithms such as Adam or stochastic gradient descent (SGD). However, since limited data are available in the lettuce browning prediction task, the model is easily susceptible to over-fitting. We adopt reptile meta-learning in the training process to alleviate this problem, a typical first-order gradient-based meta-learning algorithm for few-shot learning problems \cite{nichol2018first}. Specifically, given training batch $\boldsymbol{x}_b$ in a step with model parameters $\phi$. A traditional optimizing algorithm such as Adam or SGD simply updates the parameters as
\begin{equation}
    \tilde{\phi} = \mathrm{U}(\phi),
\end{equation}
where $\tilde{\phi}$ denotes updated parameters and $\mathrm{U}$ is the optimizing algorithm. 
As shown in Fig. \ref{reptile}, in reptile the model will sample different tasks $\tau$ on $\boldsymbol{x}_b$, then, the parameters are updated as follows:
\begin{equation}
    \begin{cases}
       & \phi' = \mathrm{U}_{\tau}^k (\phi) \\
       & \tilde{\phi} = \phi + \eta(\phi'-\phi),
    \end{cases}
\end{equation}
where $k$ stands for the steps of the optimizing operator $\mathrm{U}$ like Adam or SGD, and $\eta$ denotes the updating step for the new gradients. If $k=1$, the reptile corresponds to the base optimizer. However, when $k>1$,  the update would not assemble taking a gradient step in the base optimizer. It will include the terms from second-and-higher order gradients. Thus, the reptile will converge to a minimum that is very different from the base optimizer and fits better on the few-shot learning problem.

\section{Experiments and Results}
This section presents the dataset preparation, the experiment details and regression results. The SQ-Swin is implemented on the fresh-cut lettuce dataset collected by our group. The proposed model is quantitatively compared with other state-of-the-art models. Moreover, the interpretability of the proposed SQ-Swin is derived for more reliable application. 

\begin{figure*}
\centering
\includegraphics[width=0.85\textwidth]{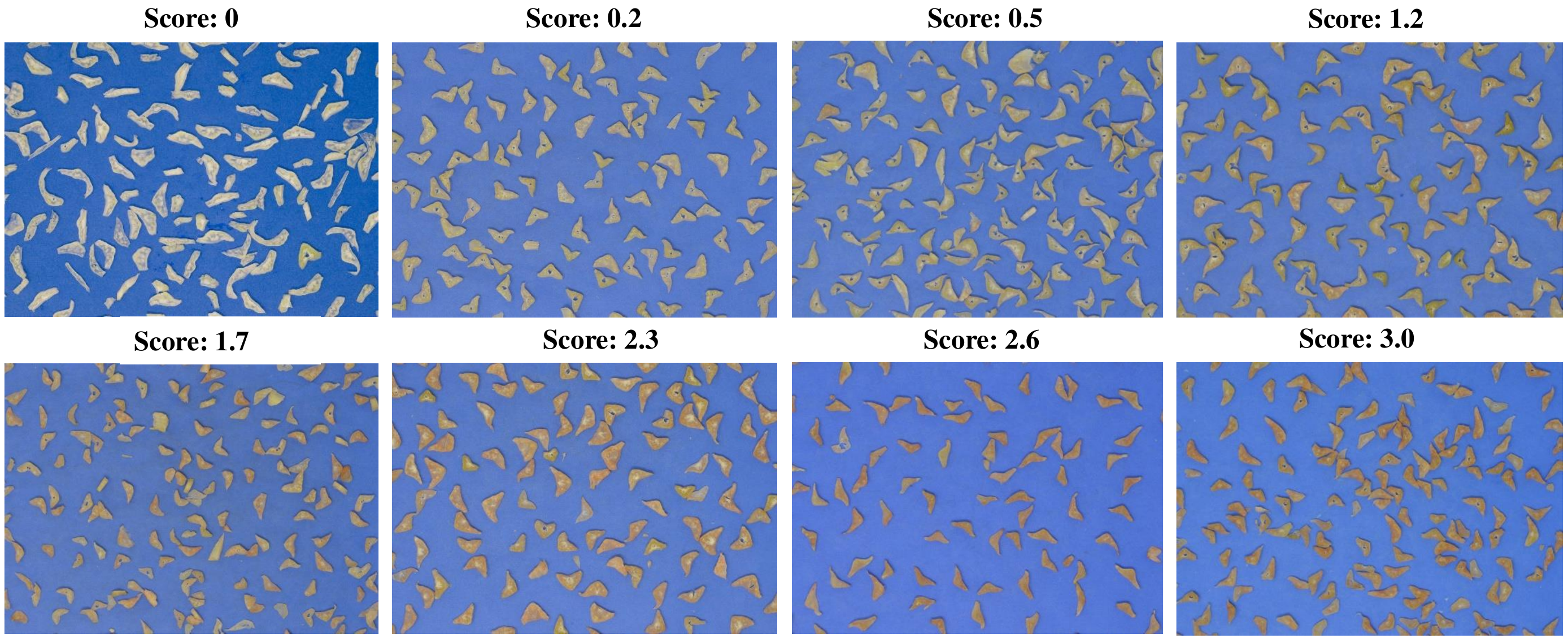}
\caption{Some fresh-cut lettuce examples with different browning scores.} 
\label{demo}
\end{figure*}

\subsection{Dataset Preparation}
Lettuce samples were grown, harvested, and processed according to Teng \textit{et. al.} (2019) \cite{teng2019identification} and Peng \textit{et. al.} (2021) \cite{peng2021phenotypic}. Digital images  of the cut lettuce were acquired inside a shooting tent with controlled lighting and shooting distance using a Nikon D1X cameras. Then, the browning level of each image is determined independently by three trained evaluators. Finally, the label of each image is an average of the three scores from the practitioners. There are 100 lettuce images of $7360\times4912$ in total with different browning levels, of which 80\% images are used for training and 20\% images for testing. The browning scores range continuously from 0 to 3.0, corresponding to different brown levels as indicated in Table \ref{brown}. Fig. \ref{demo} also shows some lettuce images with different browning scores. In the multiple scale patch extraction stage, the original patches are randomly extracted from the raw image $7360\times4912$ with size of $1024\times1024$, $2048\times2048$, $3072\times3072$, or $4096\times4096$. Forty patches are extracted from a single image. Then, all these extracted patches are resized into $224\times224$ with bilinear interpolation.
Two image transformations flipping (up/down, left/right) and rotation ($90\degree$, $180\degree$, or $270\degree$) are also applied to augment the dataset. 
\begin{table}[h]
\caption{Different browning score and description.}
    \centering
    \begin{tabular}{|c|c|}
    \hline
    Score    &  Description \\
    \hline
    0 &  fresh \\
    1.0 &  slight brown \\
    2.0 &  medium brown \\
    3.0 &  heavy brown \\
    \hline
    \end{tabular}
   \label{brown}
    \hfill
\end{table}

\subsection{Experiment Settings}
All experiments are conducted on Ubuntu 18.04.5 LTS, Intel(R) Core (TM) i9-9920X CPU @ 3.50GHz using PyTorch 1.7.1 \cite{RN102}, CUDA 10.2.0. with four NVIDIA 2080TI 11G GPUs. Below are the details of the experiment settings:
\begin{itemize}
\item We adopt the default version with {2, 2, 6, 2} layers in the four stages for the numbers of layers from the base Swin transformer modules. Between each stage, there is a downsampling and window shift operation to facilitate diverse representations.
\item The window size for each Q-Swin transformer stage is 7, and the shift is $\frac{1}{4}$ of the corresponding feature map.
\item After pretrained on the ImageNet, an MLP of size 100 is attached on top of the SQ-Swin for the feature extraction. Then, a regression head is appended for browning level prediction. 
\item For reptile meta-learning, the mini-batch size for each sampled task is 32, and the inner epoch is 4 with a learning rate of 0.6.
\item In the meta step of the reptile, the model is trained using Adam through 200 epochs with a batch size of 256. Except for the quadratic terms, all other parameters are optimized with an initial learning rate of $1\times10^{-4}$ and scheduled learning rate decay of 0.2 at epochs 100 and 150. 
\item The training of the quadratic terms utilizes the \textit{Relinear} strategy. The quadratic terms are initially set untrainable and begin to train on epoch 50. The initial learning rate of the quadratic terms are $1\times10^{-6}$, and then it is decreased to $2\times10^{-7}$ and $4\times10^{-8}$ on epoch 100 and 150, respectively.

\end{itemize}

\subsection{Evaluation Metrics}
Three metrices: \textit{mean absolute error} (MAE), \textit{mean square error} (MSE) and \textit{Pearson’s correlation coefficient} (PCC) \cite{benesty2009pearson} are employed to measure the prediction performance quantitatively. Given the true labels $\boldsymbol{y}$ and the model predictions $\hat{\boldsymbol{y}}$, MAE is expressed as $\mathrm{MAE}(\boldsymbol{y},\hat{\boldsymbol{y}}) = \frac{1}{N} |  \boldsymbol{y} - \hat{\boldsymbol{y}} |$, where $N$ is the number of testing samples.
It is a straightforward representation of the prediction errors, thereby more intuitive. Compared with other metrics, it provides more insight and interpretation of the result. 
Furthermore, to better evaluate the performance on the outlier data, MSE is also adopted $\mathrm{MSE}(\boldsymbol{y},\hat{\boldsymbol{y}}) = \frac{1}{N}|| \boldsymbol{y} - \hat{\boldsymbol{y}} ||^2$, since it tends to penalize more significant prediction errors than smaller one, thus magnifying or inflating the mean error. Next, PCC is employed to measure the association or the statistical relationship between the prediction and the truth to make the prediction more reliable. Using covariance in the formation, it informs us the linear correlation between the two variables.

\begin{equation}
    \mathrm{PCC}(\boldsymbol{y},\hat{\boldsymbol{y}}) = \frac{Cov(\boldsymbol{y},\hat{\boldsymbol{y}})} {\sigma_{\boldsymbol{y}} \sigma_{\hat{\boldsymbol{y}}}}, 
\end{equation}
where $Cov(\cdot)$ indicates the covariance. $\sigma_{\boldsymbol{y}}$ and $\sigma_{\hat{\boldsymbol{y}}}$ are the standard deviations.

\subsection{Comparative Analysis}
The SQ-Swin is fairly compared with the state-of-the-art deep convolution neural network and transformer models. The convolution models include VGG11 \cite{simonyan2014very}, Ghostnet \cite{han2020ghostnet}, and Resnet18 \cite{he2016deep}. The transformer models include Twins \cite{chu2021twins}, Deit \cite{touvron2021training}, TNT \cite{han2021transformer}, Visformer \cite{chen2021visformer}, and Swin transformer \cite{RN74}. We reimplement these models on the lettuce data according to their officially disclosed codes. Specifically, the initial parameters of these models are transfered from the pretrained models in \textit{Pytorch Image Models} \cite{rw2019timm}. Then, we add a feature extraction layer and a regression head to these models to match the lettuce browning prediction task.


\begin{figure}
\centering
\includegraphics[width=\linewidth]{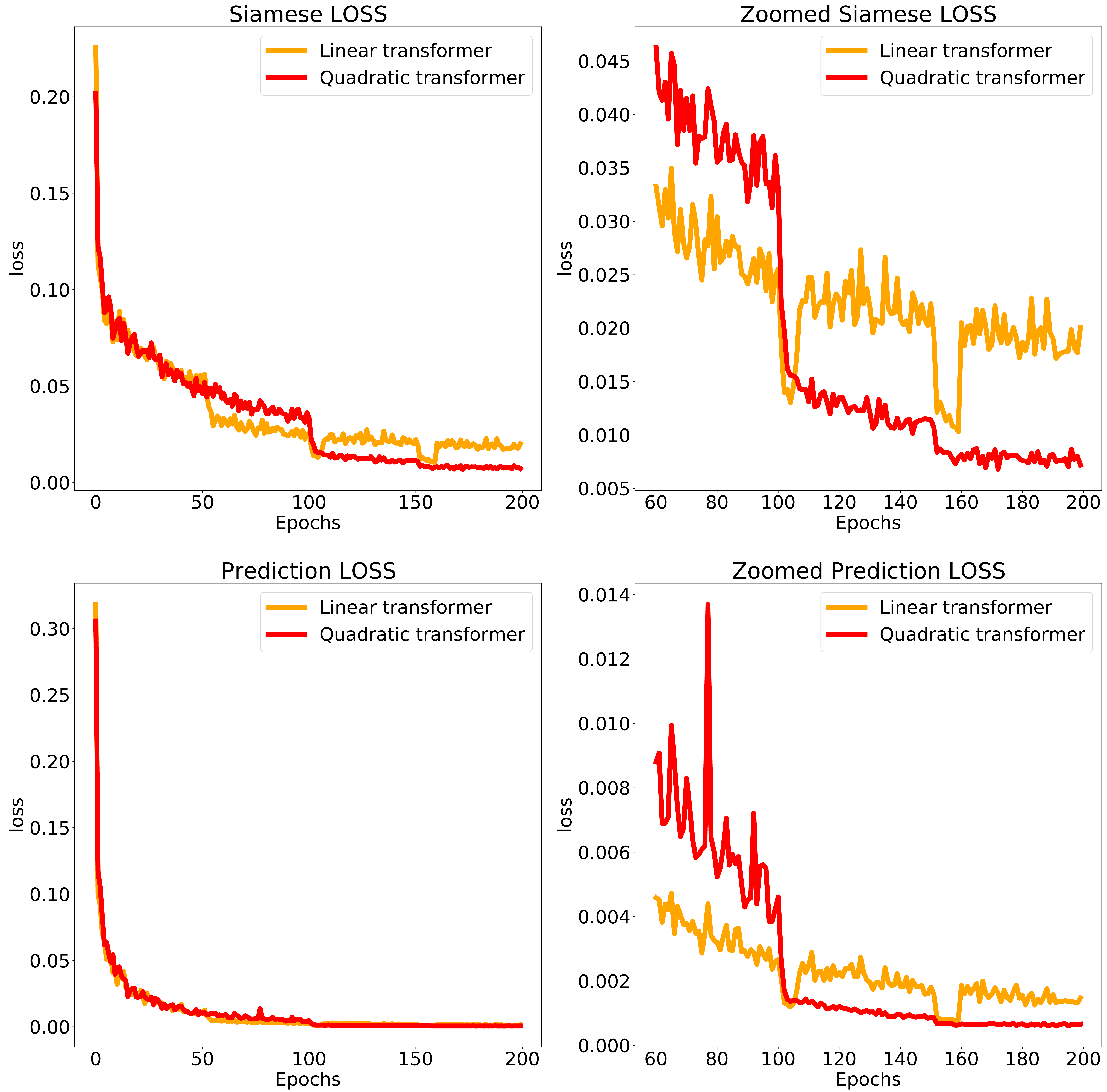}
\caption{The siamese and prediction losses of the linear transformer and the quadratic transformer.} 
\label{loss}
\end{figure}

\begin{table}[h]
\caption{Quantitative evaluation results of different methods on the testing data. The proposed model is marked by a star, and the best results are bold-font.}
    \centering
    \begin{tabular}{|l|c|c|c|c|}
    \hline
    Model    &   Type &  MAE$\downarrow$ &   MSE$\downarrow$ & PCC$\uparrow$ \\
    \hline
    vgg11     & Conv  &  0.1165 & 0.0261 & 0.9696 \\
    Densenet121 & Conv  &  0.1177 & 0.0247 & 0.9710 \\
    Resnet18 & Conv &  0.1210 & 0.0265 & 0.9691\\
    Res2net  & Conv    &  0.1349 & 0.0305 & 0.9626\\
    \hline
    Twins& Trans  & 0.1384 & 0.0328 & 0.9580\\
    Deit & Trans&   0.1231 & 0.0269 & 0.9674\\
    TNT & Trans&  0.1260 & 0.0276 & 0.9648 \\
    Visformer& Trans  & 0.1338 & 0.0300 & 0.9628\\
    Swin-T & Trans    &  0.1253 & 0.0269 & 0.9661\\
    \hline
    S-Swin* & Trans    &  0.0280 &  0.0027 & 0.9968\\
    SQ-Swin* & Trans &  \textbf{0.0203} &  \textbf{0.0018} & \textbf{0.9976}\\
    \hline
    \end{tabular}
   \label{quant}
    \hfill
\end{table}

\textbf{Quantitative Results.} 
Table \ref{quant} shows that all studied models are effective in predicting the lettuce browning level. All other convolution or transformer-based competitors can achieve a low MAE around 0.12, an MSE around 0.02, and a PCC around 0.96. However, with the siamese and reptile design, the proposed model greatly surpasses other state-of-the-art deep models in terms of the three metrics, even utilizing a traditional linear swin transformer as the base network. When the quadratic transformer is introduced, the performance further surpasses the traditional linear one with an MAE of 0.0203, an MSE of 0.0018, and a PCC of 0.9976.

\begin{figure*}
\centering
\includegraphics[width=\linewidth]{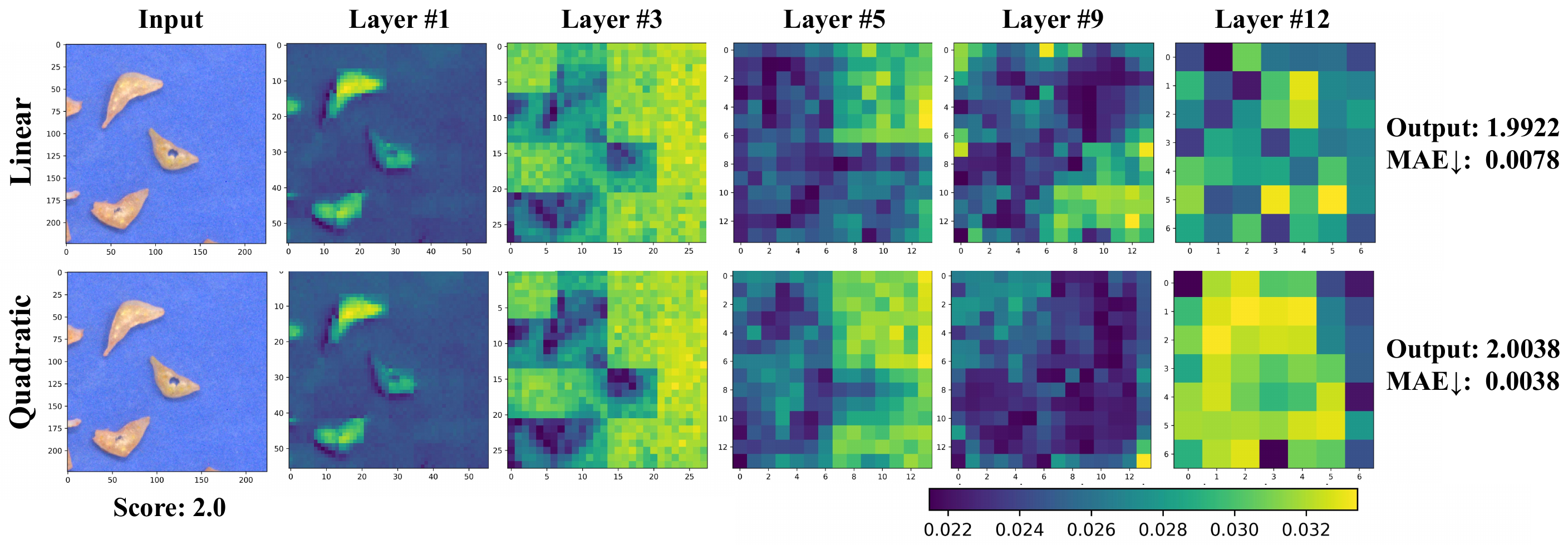}
\caption{Representative feature maps of the linear transformer and the quadratic transformer.} 
\label{feature}
\end{figure*}

\textbf{Quadratic vs. Linear Transformer.}
Besides, we also illustrate the prediction losses and the siamese losses during the training of the quadratic (SQ-Swin) and linear (S-Swin) transformers. As shown in Fig. \ref{loss}, both losses of the two transformers drops in the same way in epoch 0-50, because quadratic terms are frozen at the start. Then, the loss of the linear transformer drops faster than the loss of the quadratic transformer in epoch 50-100. However, when the learning rate decreases in epoch 100 and 150, the quadratic loss drops more while the linear loss drops but rise again. The zoomed curves of epoch 60-100 also show that the quadratic curve is smoother than the linear one. This further indicates a more stable training process. 

After training, as shown in Fig. \ref{feature}, the feature maps of the quadratic transformer and the linear transformer are also visualized. In the early layers, the two transformers follow very similar learning patterns. The feature outline and texture are nearly the same. However, when the networks go deep, the features gradually become distinct. The last layer indicates that the quadratic transformer can learn more unified and polarized features, which to some extent, provides more representative features for the browning evaluation, thereby obtaining a more accurate prediction with lower MAE. In summary, the quadratic transformer is more powerful in loss optimization, training stability, and feature representations compared with the linear transformer.

\textbf{Inference Time and Parameters.}
The traditional lettuce image analysis uses Progenesis QI software (Nonlinear Dynamics, Newcastle, U.K.) for tabular data extraction. It takes about 0.5-16 minutes of inference time, and then it uses principal component analysis (PCA) to extract principal components \cite{liu2021identification}. The whole process is rather time-consuming, thus hindering its real-world applications. Therefore, we are motivated to study the inference time of the proposed SQ-Swin on a single image. During inference, the SQ-Swin evaluates 40 randomly selected multi-scale patches from each testing image of $7360 \times 4912$, then the averaged score is used as the prediction. Table \ref{inference} shows that the inference time of the two deep learning methods is within one second which is much shorter than that of the traditional model. Moreover, the employment of the siamese architecture further expedites the inference process since the siamese models can process two images simultaneously in theory. 

Further, The number of the parameters and multiply-accumulate operations (MACs) of the SQ-Swin are also studied. As shown in Table \ref{parameter}, the SQ-Swin has few more parameters because the quadratic mapping introduces extra computation. But the computation cost is still comparable to that of the linear transformer (S-Swin). Please note that the MACs are fewer in SQ-Swin because we rewrote part of the inherent torch code. 

\begin{table}[h]
\caption{Inference time of the traditional model, Densenet121, and SQ-Swin on a single lettuce image.}
    \centering
    \begin{tabular}{|c|c|c|}
    \hline
    Model    &  Type  & Time  \\
    \hline
    Progenesis QI & Tradition & 0.5-16min  \\
    Densenet121 & Non-Siamese (deep learning) & 0.85s   \\
    SQ-Swin & Siamese (deep learning)  & 0.62s   \\
    \hline
    \end{tabular}
   \label{inference}
    \hfill
\end{table}

\begin{table}[h]
\caption{Parameters of the SQ-Swin and S-Swin.}
    \centering
    \begin{tabular}{|c|c|c|}
    \hline
    Model    &  \#param.  &  MACs \\
    \hline
    S-Swin & 28.39M & 4.36G \\
    SQ-Swin &  41.36M  &  3.32G \\
    \hline
    \end{tabular}
   \label{parameter}
    \hfill
\end{table}

\subsection{Visual Interpretability}
In real-world applications, interpretability is a desired property of any proposed model. The higher interpretability of a model is, the more people will comprehend how the decisions are made. Therefore, to further boost the reliability and the applicability of the SQ-Swin, we derive some visual interpretation of the model by leveraging the inherent attention mechanism in the model. Specifically, we extract the attention map $\mathrm{Att}(\mathbf{Q},\mathbf{K}) = \mathrm{softmax}(\frac{\mathbf{Q}\mathbf{K^\top}}{\sqrt{d_k}})$ from the multiple head attention module and analyze the internal patterns of the attention maps. 

\begin{figure}
\centering
\includegraphics[width=.95\linewidth]{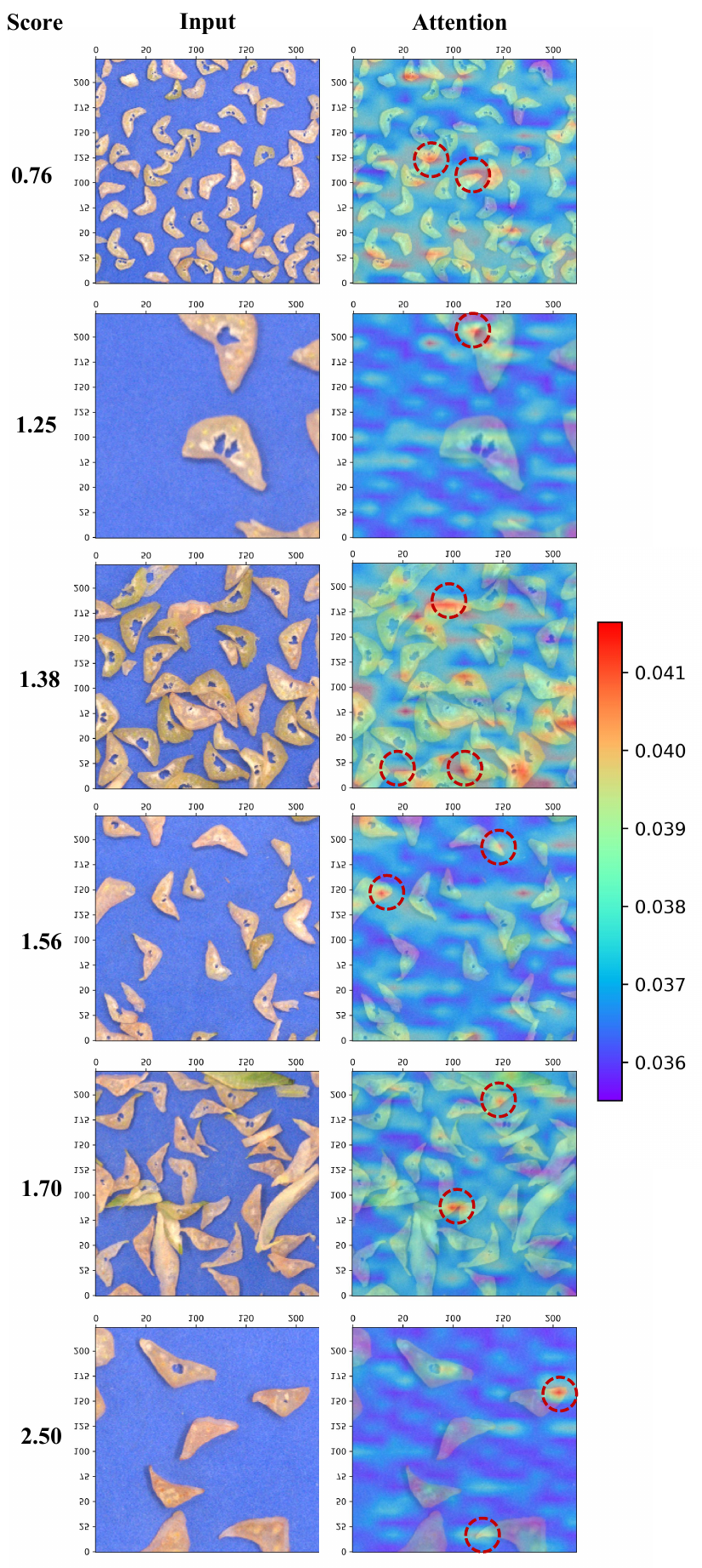}
\caption{Examples of the attention maps for different lettuce patches.} 
\label{attention}
\end{figure}

By visualizing the initial attention map of the pretrained SQ-Swin in Fig. \ref{attention}, it is noticed that the attention basically focuses on the desired fresh-cut lettuce pieces. Moreover, by investigating the lettuce parts on which the attention centers as indicated by the red dotted circle in Fig. \ref{attention}, it illustrates that the attention tends to focus on the corner or the boundary of the lettuce pieces. Therefore, we conclude that the SQ-Swin somehow relies on the corner or the boundary part of the lettuce pieces to make a browning prediction decision. This corresponds to the fact that the browning process on a fresh-cut vegetable piece usually starts from the corners and boundaries.  

\subsection{Ablation Study}
In the SQ-Swin transformer, several important methods are employed to boost the model performance: multi-scale patch,  pretraining, siamese architecture, reptile meta-learning, and quadratic transformer. In this part, we intensively investigate the impacts of these methods. MAE is utilized to quantitatively evaluate the prediction performances since it informs us the intuitive gap between the prediction and the truth. Table \ref{ablation} indicates that all studied methods can effectively boost the model performances. For example, adopting the pretraining, siamese architecture, and the reptile learning method improves the performance with an MAE above 0.4. Following up, the multi-scale patch also decreases the prediction error by 0.0337. However, the performance of the quadratic transformer further surpasses the excellent scores under these four methods with an improvement of 0.0077. In conclusion, all investigated methods  in the SQ-Swin are valid and effective for the lettuce browning prediction problem. 
\begin{table}[h]
\caption{Quantitative evaluation results of different methods on the testing data. The bold-font text indicates the new method added to the model in the previous row. The bold-font numbers indicate the best results. }
    \centering
    \begin{tabular}{|l|c|c|c|c|c|}
    \hline
    SQ-Swin    &  MAE$\downarrow$ &  Boost  \\
    \hline
    none & 0.1921 &  - \\
    \textbf{multi-scale} & 0.1584 &  0.0337$\downarrow$ \\
    multi-scale+\textbf{pretrain} & 0.1174 &  0.0410$\downarrow$ \\
    multi-scale+pretrained+\textbf{siamese} & 0.0712 &  0.0462$\downarrow$  \\
    multi-scale+pretrained+siamese+\textbf{reptile}  &  0.0280 &  0.0432$\downarrow$ \\
    \hline
    multi-scale+pretrained+siamese+reptile+\textbf{quadratic} &   \textbf{0.0203} &   0.0077$\downarrow$ \\
    \hline
    \end{tabular}
   \label{ablation}
    \hfill
\end{table}

\section{Conclusion}
In this paper, we propose a pretrained siamese quadratic swin transformer for lettuce browning prediction. To the best of our knowledge, this is the first-of-its-kind deep learning model in the lettuce brown detection field, and it greatly improves the inference time for the lettuce browning prediction. Moreover, compared with a typical linear transformer, the proposed SQ-Swin is the first second-order transformer that is more powerful in capturing real-world representations with comparable computation complexity. Experimental results validate its advantages in loss optimization, training stability, and feature representations. Furthermore, we unveil the visual interpretability of the SQ-Swin for browning prediction, which facilitates its reliability and applicability. In addition, comprehensive ablation studies show that all components, including multiscale patch, pretraining, siamese architecture, reptile meta-learning, and quadratic transformer, are effective in the lettuce browning prediction. Finally, extensive comparative experiments indicate that the SQ-Swin outperforms the state-of-the-art methods in terms of MAE, MSE, and PCC. In the future, we will explore further the application of the SQ-Swin to other food science areas.  

\section{Acknowledgement}
Authors wish to thank Drs. Ivan Simko and Hui Peng at USDA-ARS in Salinas CA for growing and harvesting lettuce, and Drs. Bin Zhou, Zi Teng, Ms. Ellen Turner, and Mr. Daniel Pearlstein at USDA-ARS  in Beltsville MD for processing and evaluating lettuce.  

\bibliographystyle{ieeetr}
\bibliography{lettuce.bib}
\end{document}